\documentclass[letterpaper,10pt,conference]{format/ieeeconf}

\IEEEoverridecommandlockouts
\usepackage{graphicx}
\usepackage{booktabs}
\usepackage{cite}
\usepackage{url}
\usepackage{balance}
\usepackage{tikz}

\graphicspath{ {./figures/} }

\newcommand\copyrighttext{%
  \scriptsize\centering
  \textcopyright\ 2026 IEEE. Personal use of this material is permitted.
  Permission from IEEE must be obtained for all other uses, in any current or future
  media, including reprinting/republishing this material for advertising or promotional
  purposes, creating new collective works, for resale or redistribution to servers or
  lists, or reuse of any copyrighted component of this work in other works.\\
  Accepted manuscript. Accepted for publication in the 2026 IEEE National Aerospace
  and Electronics Conference (NAECON). DOI not yet assigned.}
\newcommand\copyrightnotice{%
\enlargethispage{-0.7in}%
\begin{tikzpicture}[remember picture,overlay]
\node[anchor=south,yshift=10pt] at (current page.south) {\fbox{\parbox{\dimexpr\textwidth-\fboxsep-\fboxrule\relax}{\copyrighttext}}};
\end{tikzpicture}%
}

\title{\LARGE \bf
Learning to Stack: Cube-Stacking Imitation Learning from Virtual Reality Demonstrations
}

\author{Gryffin Reizian$^{1^{\dagger}}$, Jordan Dowdy$^{1^{*\dagger}}$and Jean Chagas Vaz$^{2}$
\thanks{$^{1\dagger}$G. Reizian, an undergraduate student, and $^{1^{*\dagger}}$J. Dowdy, a PhD student, are both with the Department of Electrical and Computer Engineering, University of Louisville. {\tt\small jordan.dowdy@louisville.edu}}
\thanks{$^{2}$Dr. Jean Chagas Vaz is with the Faculty of Electrical and Computer Engineering at the University of Louisville, Louisville, KY 40208, USA. {\tt\small jean.chagasvaz@louisville.edu}}
\thanks{$^{\dagger}$these authors had equal contribution.}
\thanks{$^{*}$direct all correspondence to this author.}
}

\begin{document}

\maketitle
\thispagestyle{empty}
\pagestyle{empty}
\copyrightnotice

\begin{abstract}
Imitation learning is attractive for robot manipulation, but collecting demonstrations remains a bottleneck for multi-stage tasks requiring repeated scene resets. This work presents a virtual-reality data-collection pipeline for cube-stacking with a custom 5-DoF arm in NVIDIA Isaac Sim and Isaac Lab. Using an HTC Vive Pro 2, Manus Quantum gloves, and OpenXR, an operator provides SE(3) end-effector commands to generate task demonstrations. The proposed framework separates demonstration collection from dataset construction by replaying recorded trajectories, converting task-space commands into joint-space actions, and re-rendering demonstrations with updated sensor or state configurations. This allows previously collected demonstrations to be reused for new observation and action spaces without repeating human teleoperation. The task requires stacking the red cube on the blue cube and the green cube on the red cube, with randomized cube placement. In 30 minutes, 200 virtual demonstrations were collected, compared with 45 real-world demonstrations, and Isaac Mimic generated 100 additional samples. A behavior-cloning policy was trained from the virtual demonstrations using LeRobot-style dual-camera observations and evaluated in simulation.
\end{abstract}

\section{INTRODUCTION}
Imitation learning has become a practical route for robot manipulation because policies can be learned directly from expert behavior rather than hand-designed logic~\cite{Ross2011DAgger}. Recent progress has been accelerated by better teleoperation interfaces, open datasets, and shared software stacks for training and evaluation~\cite{robomimic2021,ALOHAUnleashed2024,LeRobot2026}. Such technology has the potential to expand robot capabilities in manufacturing, logistics, and other settings where repeated manipulation tasks are required. Despite the enormous potential of imitation learning, data remains a key bottleneck. Long-horizon demonstrations require repeated object resets, robot repositioning, and scene recovery after failed trials. These costs are especially apparent in cube-stacking tasks, where each trial depends on a clean initial scene.
\begin{figure}[!t]
	\centering
    \includegraphics[width=0.99\linewidth]{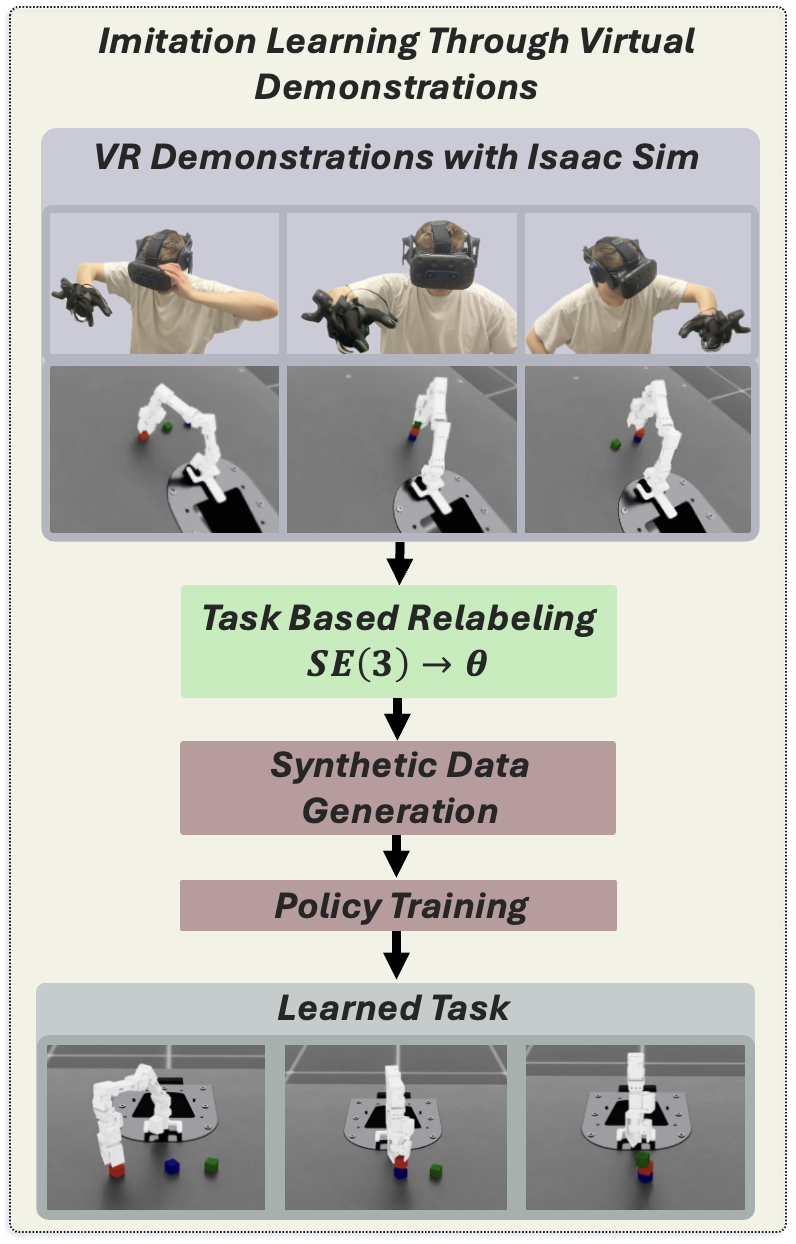}
	\caption{Overview of the proposed virtual reality imitation-learning Isaac Sim pipeline.}
    \vspace{-4mm}
	\label{fig:system_overview}
\end{figure}
Virtual environments provide a direct way to reduce these costs. In simulation, tasks can be reset instantly, demonstrations can be replayed repeatedly, and sensor configurations can be changed after collection. NVIDIA Isaac Gym and Isaac Lab have made this style of development more practical by combining accelerated physics simulation with robot-learning tooling~\cite{IsaacGym,IsaacLabMimic2026}. Isaac Lab Mimic further provides a framework for augmenting small human demonstration sets through replay, subtask annotation, and trajectory stitching~\cite{IsaacLabMimic2026}. That said, an efficient human interface is still required if the initial demonstrations are to be collected quickly and with sufficient quality.

This paper focuses on a compact manipulation platform. The robot used in this work is a custom 5-DoF arm of roughly the same scale as recently popular low-cost manipulation platforms~\cite{LeRobot2026}. Smaller manipulators lower hardware cost, reduce setup time, and make rapid iteration on teleoperation and data pipelines more accessible to research groups. That said, such systems are still capable of producing important, relevant, and foundational data for future manipulation policies. The present work builds on prior lab efforts in immersive telepresence and Isaac-based simulation~\cite{Dowdy2025Humanoids,Dowdy2025IsaacSim}, but redirects those ideas toward imitation-learning data collection for a compact cube-stacking task. Therefore, a robust, reusable method for collecting virtual manipulation demonstrations with a small arm is needed.

Figure~\ref{fig:system_overview} shows the intended structure of the proposed system, from immersive teleoperation to replay validation and synthetic dataset expansion.

\subsection*{Paper Contributions}
This paper's main contributions are:
\begin{enumerate}
    \item A virtual reality teleoperation pipeline for collecting cube-stacking demonstrations for LeRobot-style robotic arms in Isaac Sim and Isaac Lab.
    \item A replay and retargeting process that verifies demonstrations, converts SE(3) inverse-kinematics actions into joint-space actions, and supports later sensor or state re-rendering for model training.
    \item A data-efficiency comparison between virtual and real-world demonstration collection, with additional Isaac Mimic dataset expansion.
    \item Policy evaluation through Isaac Sim, with dual-camera observation, vision domain randomization, and a ResNet18-LSTM imitation-learning policy.
\end{enumerate}

\subsection*{Article Structure}
This paper contains six sections: Section II reviews related work in teleoperation, imitation learning, and Isaac-based data generation. Section III details the proposed virtual demonstration pipeline. Section IV explains the experimental setup, while Section V discusses the results and practical implications. Lastly, Section VI concludes the paper and comments on future work.

\section{RELATED WORK}
VR telemanipulation, imitation learning, and simulation-based data generation are topics that the robotics community has previously explored. The following subsections present several relevant works from the literature.

\subsubsection{VR Teleoperation and Telepresence}
Immersive teleoperation has become an important tool for collecting robot demonstrations because it allows a human operator to provide task intent directly through motion. RoboTurk showed that remote teleoperation can scale manipulation demonstrations for imitation learning~\cite{Mandlekar2018RoboTurk}, while prior telepresence work showed that avatar-style control can improve remote manipulation and embodied operation for humanoid systems~\cite{Vaz2024}. More recently, Open-TeleVision used immersive active visual feedback to collect long-horizon manipulation data for robot learning~\cite{OpenTeleVision2025}. Our own recent miniature humanoid tele-loco-manipulation work also showed that virtual reality can provide an intuitive control interface for small embodied platforms~\cite{Dowdy2025Humanoids}. That said, these works were not centered on rapid dataset collection for a fixed-base manipulation task in simulation.

\subsubsection{Imitation Learning and Shared Demonstration Pipelines}
Learning from demonstrations has benefited from more reproducible datasets and open tooling. Robomimic highlighted the importance of dataset quality and algorithm choice when learning from offline human manipulation demonstrations~\cite{robomimic2021}. ALOHA Unleashed showed that scaling data collection on accessible hardware can improve imitation-learning performance on dexterous tasks~\cite{ALOHAUnleashed2024}. Likewise, LeRobot reflects the trend toward shared low-cost hardware, common data formats, and end-to-end robot-learning infrastructure~\cite{LeRobot2026}. Deep visuomotor policy work also motivates learning policies directly from camera observations for manipulation~\cite{Levine2016Visuomotor}. These works motivate efficient demonstration pipelines that reduce operator burden while preserving data quality, especially for long-horizon manipulation tasks.

\subsubsection{Isaac-Based Data Generation}
Accelerated simulation has become central to modern robot learning. Isaac Gym demonstrated that massively parallel GPU physics can support rapid robot-learning workflows~\cite{IsaacGym}. Isaac Lab extends this with a broader development framework for reinforcement learning, imitation learning, and robot task design in Isaac Sim. Of particular relevance here, Isaac Lab Mimic formalizes a pipeline where demonstrations are recorded, annotated into subtasks, replayed, transformed relative to task objects, and used to generate additional trajectories for downstream Robomimic behavior cloning~\cite{IsaacLabMimic2026}. Beyond fixed-base manipulation, demonstration-driven embodied learning has also been explored for visual loco-manipulation~\cite{2403.20328}. The present work adopts this general philosophy, but applies it to a custom small arm, immersive VR teleoperation, and a cube-stacking task where data-collection efficiency is the main concern.

\section{Methodology}
This section outlines the proposed virtual reality pipeline for generating cube-stacking demonstrations and preparing them for imitation-learning experiments. Additional implementation details are provided for the VR interface, replay process, and synthetic data generation.

\subsection{System Overview}
The manipulation environment was developed in NVIDIA Isaac Sim and Isaac Lab 2.3.2. A custom 5-DoF arm was integrated into the simulator as the robot embodiment used for both teleoperation and replay. The small scale of the arm makes the platform inexpensive, easy to iterate on, and representative of the kind of benchtop hardware that many research groups can practically deploy.

Telemanipulation control in VR is realized using an HTC Vive Pro 2 headset and Manus Quantum gloves. OpenXR serves as the translation layer between VR hardware and the simulation stack~\cite{OpenXR2024}. During collection, the operator commands the robot end effector directly in SE(3), while a numerical Pink inverse kinematics solver maps these task-space targets to joint motion for the custom arm~\cite{Pink2025}. This control choice was made because task-space teleoperation is more intuitive for grasping and placement than direct joint-level control. Notwithstanding the compact arm morphology, the same approach can be generalized to other small manipulation platforms.

\subsection{Cube-Stacking Task and Observations}
The task environment is a sequential cube-stacking problem: the robot first places the red cube on the blue cube, then places the green cube on the red cube. Cube positions are randomized at the beginning of each demonstration, so the dataset does not collapse to a single nominal trajectory.

Two RGB camera views are used to generate policy-training data. These observations are added during replay rather than fixed during the original VR demonstration, allowing the same trajectory to be re-rendered with updated sensors, model-state observations, or actions. One camera is fixed approximately one meter behind the robot and views the arm and three cubes, while the other is mounted near the gripper and points toward the pinch region. During replay and training-data export, the background skybox, table, and block textures, reflectivity, material type, and brightness are randomized to reduce dependence on one fixed scene appearance, following the general idea of visual domain randomization~\cite{Tobin2017DomainRandomization}.

\begin{figure}[!t]
    \centering
    \vspace{3mm}

    \setlength{\fboxsep}{-0.0pt}%
    \setlength{\fboxrule}{2.0pt}%
    \fbox{%
        \includegraphics[width=\dimexpr\linewidth-2\fboxrule\relax]{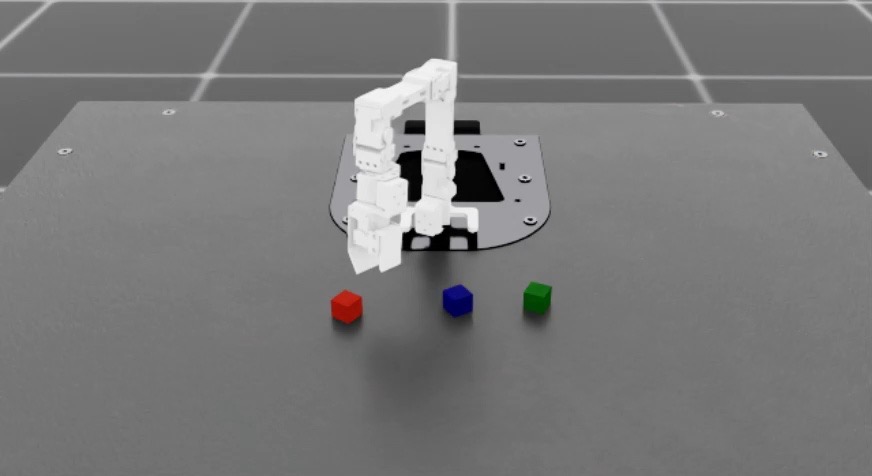}%
    }

    \caption{Cube-stacking task environment used for data collection with a custom 5-DoF arm, where the red, blue, and green cubes must be stacked in order, respectively.}
    \label{fig:task_env}
    \vspace{-4mm}
\end{figure}
Figure~\ref{fig:task_env} shows the intended scene layout, including the cube-stacking workspace used to generate the dataset observations.

\subsection{Replay, Retargeting, and Dataset Generation}
The pipeline separates human teleoperation, deterministic replay, and final dataset export. During collection, demonstrations store task-space end-effector motion and scene evolution. Each trajectory is then replayed to verify reproducibility and converted from inverse-kinematics pose actions into joint-space position sequences.

This replay structure makes the demonstrations reusable: sensor placement, model-state observations, and action representations can be changed after collection without discarding the original operator data. To further expand the dataset, Isaac Mimic is used to split demonstrations into subtasks, transform them relative to object poses, stitch new trajectories under randomized object initializations, and filter by task success before export~\cite{IsaacLabMimic2026}. The resulting data remains compatible with downstream Robomimic behavior-cloning workflows~\cite{robomimic2021}.

\begin{figure}[!t]
    \centering
    \vspace{3mm}
        \includegraphics[width=0.98\linewidth]{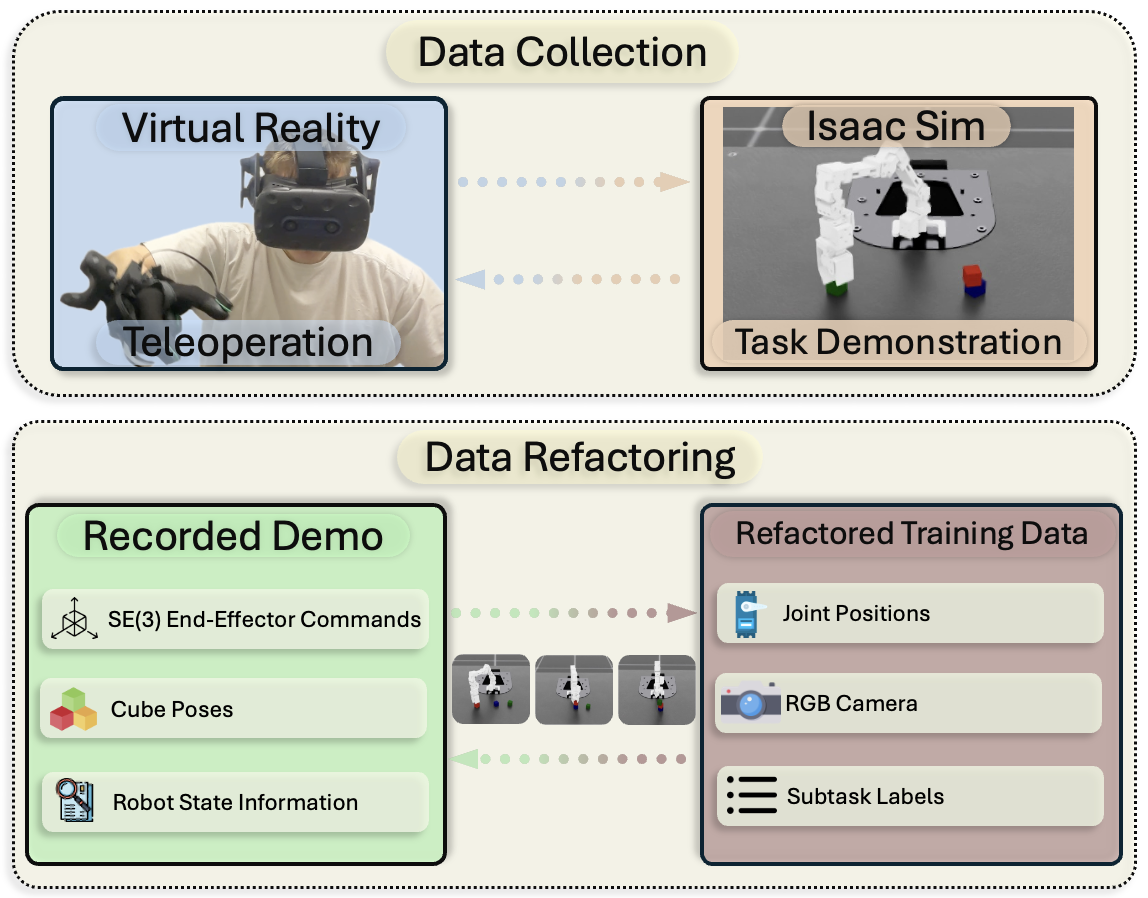}%
    \caption{Robot state retargeting, data replay, and synthetic data generation workflow.}
    \label{fig:replay_pipeline}
    \vspace{-4mm}
\end{figure}

Figure~\ref{fig:replay_pipeline} summarizes how the collected demonstrations are converted into reusable joint-space trajectories and additional synthetic samples.

\section{EXPERIMENTATION}
This section evaluates the pipeline through demonstration throughput, dataset expansion, and an initial simulation-only policy test.

\subsection{Manual Demonstration Collection}
A human operator used the VR interface to collect cube-stacking demonstrations in simulation. Over a $30$-minute session, approximately $200$ virtual demonstrations were collected. For comparison, a real-world workflow required manual resets and achieved approximately $45.0$ demonstrations in the same $30$-minute period. In rate form, this corresponds to approximately $6.67$ demonstrations per minute for the virtual workflow and $1.5$ demonstrations per minute for the real-world workflow.

\subsection{Synthetic Demonstration Generation}
The $200$ manually recorded demonstrations were then used as the source set for Isaac Mimic. Over approximately three days of computing, Isaac Mimic generated $100$ additional synthetic demonstrations. This brought the total dataset size to $300$ demonstrations. While synthetic generation was slower in wall-clock time than manual VR teleoperation, it ran without continuous operator involvement, which is an important distinction for practical dataset scaling.

\subsection{Policy Training and Simulation Evaluation}
A behavior cloning policy was trained on the virtual demonstration dataset for the cube-stacking task. The replay pipeline was used to add two RGB camera observations after the original demonstrations were collected: a gripper-mounted camera and a fixed-scene camera located approximately 1 meter behind the arm, looking at the robot and all three cubes. The policy used ResNet18 visual encoders~\cite{He2016ResNet} with an LSTM temporal model~\cite{Hochreiter1997LSTM}. During training, domain randomization was applied to the environment skybox, table, and block textures, reflectivity, material type, and brightness.

The trained policy was evaluated in $30$ simulation trials with randomized block positions. Full task success requires placing the red cube on the blue cube, then placing the green cube on the red cube. Subtask success was also measured for four sequential milestones: picking up the red block, placing the red block on the blue block, picking up the green block, and placing the green block on the red block. The full task success rate was $10\%$, while the subtask success rates are summarized in Table~\ref{tab:policySummary}.

\begin{table}[!t]
\centering
\vspace{2mm}
\caption{Simulation Policy Evaluation Summary}\label{tab:policySummary}
\setlength{\tabcolsep}{4pt}
\renewcommand{\arraystretch}{1.25}
\begin{tabular}{lcc}
\toprule
\textbf{Evaluation Metric} & \textbf{Successes} & \textbf{Rate} \\
\midrule
Full Task Completion & 3/30 & 10\% \\
Subtask 1: Pickup red & 6/30 & 20\% \\
Subtask 2: Place red on blue & 6/30 & 20\% \\
Subtask 3: Pickup green & 3/30 & 10\% \\
Subtask 4: Place green on red & 3/30 & 10\% \\
\bottomrule
\end{tabular}
\vspace{-1mm}
\end{table}

\section{RESULTS AND DISCUSSION}
The proposed workflow improved demonstration throughput while keeping the dataset reusable for later policy-training experiments. Collecting $200$ virtual demonstrations in $30$ minutes, compared with $45$ real-world demonstrations in the same time, gave an approximately $4.4$$\times$ increase in demonstrations per unit operator time. This improvement mainly comes from instant scene resets in simulation and the removal of repeated manual task-state recovery.

\begin{table}\centering
\vspace{3mm}
\caption{Demonstration Collection Summary}\label{tab:dataSummary}
\setlength{\tabcolsep}{4pt}
\renewcommand{\arraystretch}{1.35}
\begin{tabular}{lccc}
\toprule
\textbf{Workflow} & \textbf{Demonstrations} & \textbf{Time} & \textbf{Rate} \\
\midrule
Virtual (Manual) & 200 & 30 min & 6.67 demo/min. \\
Real-world (Manual) & 45 & 30 min & 1.50 demo/min. \\
Isaac Mimic (Generated) & 100 & 3 days & 1.38 demo/hour \\
\bottomrule
\end{tabular}
\vspace{-1mm}
\end{table}

Table~\ref{tab:dataSummary} summarizes the collection rates. The replay-centered structure shown in Figure~\ref{fig:task_env} also allows the same demonstrations to support multiple camera or observation variants, reducing the risk of re-collecting data when the policy input design changes. Isaac Mimic added $100$ demonstrations over approximately three days; although slower in wall-clock time, this expansion was unattended and therefore complements manual VR collection.

The simulation policy evaluation shows that the replayed virtual demonstrations were usable for training, but the autonomous baseline remains limited. The policy completed the full stack in $3$ of $30$ randomized trials, giving a $10\%$ success rate. Its main failures were weak color-order disambiguation, where the policy often selected the closest cube rather than the required next cube, and gripper-alignment errors that caused cubes to jam in the pinch region.

These results support the pipeline as a practical first step for generating small-arm imitation-learning data, but not yet as a robust cube-stacking policy. The policy was evaluated only in simulation, and synthetic trajectory generation still depends on seed-demonstration quality and replay consistency. Future work should focus on larger and more diverse datasets, policy and observation ablations, improved visual disambiguation, gripper-control refinement, and deployment on physical hardware.

\section{CONCLUSION}
This paper presented a virtual reality pipeline for collecting and expanding cube-stacking imitation-learning demonstrations with a custom 5-DoF arm in Isaac Sim and Isaac Lab. Using immersive SE(3) teleoperation, replay-based validation, task-space to joint-space retargeting, dual-camera observation generation, and Isaac Mimic dataset expansion, the proposed workflow reduced operator burden while increasing demonstration throughput. Approximately $200$ virtual demonstrations were collected in $30$ minutes, compared to $45$ demonstrations for a comparable real-world process, and an additional $100$ synthetic demonstrations were generated from the $200$ manual trajectories. A preliminary dual-camera ResNet18 and LSTM policy trained from virtual demonstrations achieved $10\%$ full-task success across $30$ randomized simulation trials, with failures primarily due to confusion over cube colors and gripper alignment issues. Overall, the developed system shows potential for rapid manipulation of dataset generation on small robotic platforms. Future work will focus on improving autonomous policy reliability, expanding dataset variation, and deploying the pipeline to additional small manipulation platforms.

\balance
\bibliographystyle{./format/IEEEtran.bst}
\bibliography{./format/IEEEabrv.bib,references}

\end{document}